\documentclass[12pt]{article}

\usepackage{epsfig}

\def\be{\begin{equation}}
\def\ee{\end{equation}}
\def\la{\langle}
\def\ra{\rangle}
\def\IP{\hbox{\rm I\kern -1.6pt{\rm P}}}
\def\IC{{\hbox{\rm C\kern-.58em{\raise.53ex\hbox{$\scriptscriptstyle|$}}
    \kern-.55em{\raise.53ex\hbox{$\scriptscriptstyle|$}} }}}
\def\IN{\hbox{I\kern-.2em\hbox{N}}}
\def\IR{\hbox{\rm I\kern-.2em\hbox{\rm R}}}
\def\ZZ{\hbox{{\rm Z}\kern-.3em{\rm Z}}}
\def\IT{\hbox{\rm T\kern-.38em{\raise.415ex\hbox{$\scriptstyle|$}} }}

\begin{document}

\title{Statistical efficiency of curve fitting algorithms}
\author{N. Chernov and C. Lesort\\
Department of Mathematics\\
University of Alabama at Birmingham\\
Birmingham, AL 35294, USA}
\date{\today}
\maketitle

\begin{abstract}
We study the problem of fitting parametrized curves to noisy data.
Under certain assumptions (known as Cartesian and radial
functional models), we derive asymptotic expressions for the bias
and the covariance matrix of the parameter estimates. We also
extend Kanatani's version of the Cramer-Rao lower bound, which he
proved for unbiased estimates only, to more general estimates that
include many popular algorithms (most notably, the orthogonal
least squares and algebraic fits). We then show that the
gradient-weighted algebraic fit is statistically efficient and
describe all other statistically efficient algebraic fits.
\end{abstract}

\begin{center}
\end{center}

\renewcommand{\theequation}{\arabic{section}.\arabic{equation}}

\section{Introduction}
\label{secI} \setcounter{equation}{0}

In many applications one fits a parametrized curve described by an
implicit equation $P(x,y;\Theta)=0$ to experimental data $(x_i,y_i)$,
$i=1,\ldots,n$. Here $\Theta$ denotes the vector of unknown parameters
to be estimated. Typically, $P$ is a polynomial in $x$ and $y$, and its
coefficients are unknown parameters (or functions of unknown
parameters). For example, a number of recent publications
\cite{ARW01,CBH01,GGS94,LM00,Sp97} are devoted to the problem of
fitting quadrics $Ax^2+ Bxy+ Cy^2+ Dx+ Ey+ F=0$, in which case
$\Theta=(A,B,C,D,E,F)$ is the parameter vector. The problem of fitting
circles, given by equation $(x-a)^2+ (y-b)^2 -R^2=0$ with three
parameters $a,b,R$, also attracted attention
\cite{CO84,Ka98,La87,Sp96}.

We consider here the problem of fitting general curves given by
implicit equations $P(x,y;\Theta)=0$ with $\Theta= (\theta_1, \ldots,
\theta_k)$ being the parameter vector. Our goal is to investigate
statistical properties of various fitting algorithms. We are interested
in their biasedness, covariance matrices, and the Cramer-Rao lower
bound.

First, we specify our model. We denote by $\bar{\Theta}$ the true
value of $\Theta$. Let $(\bar{x}_{i} ,\bar{y}_{i})$,
$i=1,\ldots,n$, be some points lying on the true curve
$P(x,y;\bar{\Theta})=0$. Experimentally observed data points
$(x_i, y_i)$, $i=1,\ldots,n$, are perceived as random
perturbations of the true points $(\bar{x}_{i} ,\bar{y}_{i})$. We
use notation ${\bf x}_i = ({x}_i, {y}_i)^T$ and $\bar{\bf x}_i =
({\bar x}_i,\bar{y}_i)^T$, for brevity. The random vectors ${\bf
e}_i={\bf x}_i -\bar{\bf x}_{i}$ are assumed to be independent and
have zero mean. Two specific assumptions on their probability
distribution can be made, see \cite{BC86}:
\begin{itemize} \item[] {\em Cartesian model}: Each ${\bf e}_i$
is a two-dimensional normal vector with covariance matrix $\sigma^2_i
I$, where $I$ is the identity matrix. \item[] {\em Radial model}: ${\bf
e}_i = \xi_i {\bf n}_i$ where $\xi_i$ is a normal random variable
${\cal N}(0,\sigma^2_i)$, and ${\bf n}_i$ is a unit normal vector to
the curve $P(x,y;\bar{\Theta})=0$ at the point ${\bf x}_i$.
\end{itemize} Our analysis covers both models, Cartesian and radial.
For simplicity, we assume that $\sigma^2_i=\sigma^2$ for all $i$,
but note that our results can be easily generalized to arbitrary
$\sigma_i^2>0$.

Concerning the true points $\bar{\bf x}_i$, $i=1,\ldots,n$, two
assumptions are possible. Many researchers \cite{Ch65,Ka96,Ka98}
consider them as fixed, but unknown, points on the true curve. In
this case their coordinates $(\bar{x}_{i} ,\bar{y}_{i})$ can be
treated as additional parameters of the model (nuisance
parameters). Chan \cite{Ch65} and others \cite{An81,BC86} call
this assumption a {\em functional model}. Alternatively, one can
assume that the true points $\bar{\bf x}_i$ are sampled from the
curve $P(x,y ;\bar{\Theta} )=0$ according to some probability
distribution on it. This assumption is referred to as a {\em
structural model} \cite{An81,BC86}. We only consider the
functional model here.

It is easy to verify that maximum likelihood estimation of the
parameter $\Theta$ for the functional model is given by the
orthogonal least squares fit (OLSF), which is based on
minimization of the function
\be
       {\cal F}_1(\Theta) =  \sum_{i=1}^n [d_i(\Theta)]^2
         \label{Fmain1}
\ee
where $d_i(\Theta)$ denotes the distance from the point ${\bf x}_i$ to
the curve $P(x,y;\Theta)=0$. The OLSF is the method of choice in
practice, especially when one fits simple curves such as lines and
circles. However, for more general curves the OLSF becomes intractable,
because the precise distance $d_i$ is hard to compute. For example,
when $P$ is a generic quadric (ellipse or hyperbola), the computation
of $d_i$ is equivalent to solving a polynomial equation of degree four,
and its direct solution is known to be numerically unstable, see
\cite{ARW01,GGS94} for more detail. Then one resorts to various
approximations. It is often convenient to minimize
\be
       {\cal F}_2(\Theta) =  \sum_{i=1}^n [P(x_i,y_i;\Theta)]^2
         \label{Fmain2}
\ee
instead of (\ref{Fmain1}). This method is referred to as a
(simple) {\em algebraic fit} (AF), in this case one calls
$|P(x_i,y_i;\Theta)|$ the {\em algebraic distance}
\cite{ARW01,CBH01,GGS94} from the point $(x_i,y_i)$ to the curve.
The AF is computationally cheaper than the OLSF, but its accuracy
is often unacceptable, see below.

The simple AF (\ref{Fmain2}) can be generalized to a {\em weighted
algebraic fit}, which is based on minimization of
\be
       {\cal F}_3(\Theta) =  \sum_{i=1}^n w_i\, [P(x_i,y_i;\Theta)]^2
         \label{Fmain3}
\ee
where $w_i=w(x_i,y_i;\Theta)$ are some weights, which may balance
(\ref{Fmain2}) and improve its performance. One way to define
weights $w_i$ results from a linear approximation to $d_i$:
$$
   d_i \approx \frac{|P(x_i,y_i;\Theta)|}
     {\|\nabla_{\bf x}P(x_i,y_i;\Theta)\|}
$$
where $\nabla_{\bf x}P=(\partial P/\partial x,\partial P/\partial
y)$ is the gradient vector, see \cite{Ta91}. Then one minimizes
the function
\be
       {\cal F}_4(\Theta) =  \sum_{i=1}^n \frac{[P(x_i,y_i;\Theta)]^2}
       {\|\nabla_{\bf x}P(x_i,y_i;\Theta)\|^2}
         \label{Fmain4}
\ee
This method is called the {\em gradient weighted algebraic fit} (GRAF).
It is a particular case of (\ref{Fmain3}) with $w_i = 1/ \|\nabla_{\bf
x}P(x_i,y_i;\Theta)\|^2$.

The GRAF is known since at least 1974 \cite{Tu74} and recently
became standard for polynomial curve fitting
\cite{Ta91,LM00,CBH01}. The computational cost of GRAF depends on
the function $P(x,y;\Theta)$, but, generally, the GRAF is much
faster than the OLSF. It is also known from practice that the
accuracy of GRAF is almost as good as that of the OLSF, and our
analysis below confirms this fact. The GRAF is often claimed to be
a {\em statistically optimal} weighted algebraic fit, and we will
prove this fact as well.

Not much has been published on statistical properties of the OLSF and
algebraic fits, apart from the simplest case of fitting lines and
hyperplanes \cite{Hu97}. Chan \cite{Ch65}, Berman and Culpin
\cite{BC86} investigated circle fitting by the OLSF and the simple
algebraic fit (\ref{Fmain2}) assuming the structural model. Kanatani
\cite{Ka96,Ka98} used the Cartesian functional model and considered a
general curve fitting problem. He established an analogue of the
Rao-Cramer lower bound for unbiased estimates of $\Theta$, which we
call here Kanatani-Cramer-Rao (KCR) lower bound. He also showed that
the covariance matrices of the OLSF and the GRAF attain, to the leading
order in $\sigma$, his lower bound. We note, however, that in most
cases the OLSF and algebraic fits are {\em biased} \cite{BC86,Be89},
hence the KCR lower bound, as it is derived in \cite{Ka96,Ka98}, does
not immediately apply to these methods.

In this paper we extend the KCR lower bound to biased estimates,
which include the OLSF and all weighted algebraic fits. We prove
the KCR bound for estimates satisfying the following mild
assumption:
\medskip

\noindent{\bf Precision assumption}. For precise observations (when
${\bf x}_i = \bar{\bf x}_i$ for all $1\leq i\leq n$), the estimate
$\hat{\Theta}$ is precise, i.e.
\be
  \hat{\Theta}(\bar{\bf x}_1, \ldots, \bar{\bf x}_n) = \bar{\Theta}
       \label{Tass}
\ee
It is easy to check that the OLSF and algebraic fits
(\ref{Fmain3}) satisfy this assumption. We will also show that all
unbiased estimates of $\hat{\Theta}$ satisfy (\ref{Tass}).

We then prove that the GRAF is, indeed, a statistically efficient
fit, in the sense that its covariance matrix attains, to the
leading order in $\sigma$, the KCR lower bound. On the other hand,
rather surprisingly, we find that GRAF is not the only
statistically efficient algebraic fit, and we describe all
statistically efficient algebraic fits. Finally, we show that
Kanatani's theory and our extension to it remain valid for the
radial functional model. Our conclusions are illustrated by
numerical experiments on circle fitting algorithms.

\section{Kanatani-Cramer-Rao lower bound}
\label{secKCR} \setcounter{equation}{0}

Recall that we have adopted the functional model, in which the true
points $\bar{\bf x}_i$, $1\leq i\leq n$, are fixed. This automatically
makes the sample size $n$ fixed, hence, many classical concepts of
statistics, such as consistency and asymptotic efficiency (which
require taking the limit $n\to\infty$) lose their meaning. It is
customary, in the studies of the functional model of the curve fitting
problem, to take the limit $\sigma \to 0$ instead of $n\to\infty$, cf.\
\cite{Ka96,Ka98}. This is, by the way, not unreasonable from the
practical point of view: in many experiments, $n$ is rather small and
cannot be (easily) increased, so the limit $n\to \infty$ is of little
interest. On the other hand, when the accuracy of experimental
observations is high (thus, $\sigma$ is small), the limit $\sigma\to 0$
is quite appropriate.

Now, let $\hat{\Theta}({\bf x}_1,\ldots,{\bf x}_n)$ be an arbitrary
estimate of $\Theta$ satisfying the precision assumption (\ref{Tass}).
In our analysis we will always assume that all the underlying functions
are regular (continuous, have finite derivatives, etc.), which is a
standard assumption \cite{Ka96,Ka98}.

The mean value of the estimate $\hat{\Theta}$ is
\be
    E(\hat{\Theta}) =
  \int\cdots\int \hat{\Theta}({\bf x}_1,\ldots,{\bf x}_n)
  \, \prod_{i=1}^n f({\bf x}_i)\,
  d{\bf x}_1\cdots d{\bf x}_n
    \label{ET}
\ee
where $f({\bf x}_i)$ is the probability density function for the
random point ${\bf x}_i$, as specified by a particular model
(Cartesian or radial).

We now expand the estimate $\hat{\Theta}({\bf x}_1, \ldots, {\bf
x}_n)$ into a Taylor series about the true point $(\bar{\bf x}_1,
\ldots,
\bar{\bf x}_n)$ remembering (\ref{Tass}):
\be
   \hat{\Theta}({\bf x}_1, \ldots, {\bf x}_n) =
   \bar{\Theta} + \sum_{i=1}^n
   \Theta_i \times ({\bf x}_i - \bar{\bf x}_i)
   + {\cal O}(\sigma^2)
     \label{Texpand}
\ee
where
\be
  {\Theta}_i = \nabla_{{\bf x}_i}\hat{\Theta}
  (\bar{\bf x}_1, \ldots, \bar{\bf x}_n),
  \ \ \ \ \ i=1,\ldots,n
    \label{Ti}
\ee
and $\nabla_{{\bf x}_i}$ stands for the gradient with respect to
the variables $x_i,y_i$. In other words, $\Theta_i$ is a $k\times
2$ matrix of partial derivatives of the $k$ components of the
function $\hat{\Theta}$ with respect to the two variables $x_i$
and $y_i$, and this derivative is taken at the point $(\bar{\bf
x}_1, \ldots, \bar{\bf x}_n)$,

Substituting the expansion (\ref{Texpand}) into (\ref{ET}) gives
\be
   E(\hat{\Theta}) = \bar{\Theta} + {\cal O}(\sigma^2)
      \label{Tbias}
\ee
since $E({\bf x}_i - \bar{\bf x}_i)=0$. Hence, the bias of the
estimate $\hat{\Theta}$ is of order $\sigma^2$.

It easily follows from the expansion (\ref{Texpand}) that the
covariance matrix of the estimate $\hat{\Theta}$ is given by
$$
  {\cal C}_{\hat{\Theta}} = \sum_{i=1}^n
  \Theta_i E[({\bf x}_i - \bar{\bf x}_i)({\bf x}_i - \bar{\bf x}_i)^T]
  \Theta_i^T + {\cal O}(\sigma^4)
$$
(it is not hard to see that the cubical terms ${\cal O}(\sigma^3)$
vanish because the normal random variables with zero mean also
have zero third moment, see also \cite{Ka96}). Now, for the
Cartesian model
$$
     E[({\bf x}_i - \bar{\bf x}_i)({\bf x}_i - \bar{\bf x}_i)^T]
     =\sigma^2 I
$$
and for the radial model
$$
     E[({\bf x}_i - \bar{\bf x}_i)({\bf x}_i - \bar{\bf x}_i)^T]
     =\sigma^2 {\bf n}_i {\bf n}_i^T
$$
where ${\bf n}_i$ is a unit normal vector to the curve
$P(x,y;\bar{\Theta})=0$ at the point $\bar{\bf x}_i$. Then we obtain
\be
  {\cal C}_{\hat{\Theta}} = \sigma^2 \sum_{i=1}^n
  \Theta_i \Lambda_i \Theta_i^T + {\cal O}(\sigma^4)
     \label{Csig0}
\ee
where $\Lambda_i=I$ for the Cartesian model and $\Lambda_i={\bf n}_i
{\bf n}_i^T$ for the radial model.
\\

\noindent{\bf Lemma}. {\em We have $\Theta_i {\bf n}_i {\bf n}_i^T
\Theta_i^T = \Theta_i \Theta_i^T$ for each $i=1,\ldots,n$. Hence,
for both models, Cartesian and radial, the matrix ${\cal
C}_{\hat{\Theta}}$ is given by the same expression:}
\be
  {\cal C}_{\hat{\Theta}} = \sigma^2 \sum_{i=1}^n
  \Theta_i \Theta_i^T + {\cal O}(\sigma^4)
     \label{Csig}
\ee

This lemma is proved in Appendix.

Our next goal is now to find a lower bound for the matrix
\be
      {\cal D}_1:= \sum_{i=1}^n \Theta_i\Theta_i^T
        \label{calC1}
\ee
Following \cite{Ka96,Ka98}, we consider perturbations of the parameter
vector $\bar{\Theta} +\delta \Theta$ and the true points $\bar{\bf x}_i
+ \delta \bar{\bf x}_i$ satisfying two constraints. First, since the
true points must belong to the true curve, $P(\bar{\bf
x}_i;\bar{\Theta})=0$, we obtain, by the chain rule,
\be
   \la \nabla_{{\bf x}}\, P(\bar{\bf x}_i;\bar{\Theta}), \delta \bar{\bf x}_i \ra
   + \la \nabla_{\Theta} P(\bar{\bf x}_i;\bar{\Theta}), \delta \Theta \ra = 0
      \label{Tcon1}
\ee
where $\la \cdot, \cdot \ra$ stands for the scalar product of vectors.
Second, since the identity (\ref{Tass}) holds for all $\Theta$, we get
\be
   \sum_{i=1}^n
   \Theta_i\, \delta \bar{\bf x}_i
   = \delta \Theta
     \label{Tcon2}
\ee
by using the notation (\ref{Ti}).

Now we need to find a lower bound for the matrix (\ref{calC1})
subject to the constraints (\ref{Tcon1}) and (\ref{Tcon2}). That
bound follows from a general theorem in linear algebra:
\\

\noindent{\bf Theorem (Linear Algebra)}. {\em Let $n\geq k\geq 1$ and
$m\geq 1$. Suppose $n$ nonzero vectors $u_i\in\IR^m$ and $n$ nonzero
vectors $v_i\in\IR^k$ are given, $1\leq i\leq n$. Consider $k\times m$
matrices
$$
        X_i = \frac{v_iu_i^T}{u_i^Tu_i}\
$$
for $1\leq i\leq n$, and $k\times k$ matrix
$$
   B = \sum_{i=1}^n X_i X_i^T
   = \sum_{i=1}^n \frac{v_iv_i^T}{u_i^Tu_i}
$$
Assume that the vectors $v_1,\ldots,v_n$ span $\IR^k$ (hence $B$
is nonsingular). We say that a set of $n$ matrices
$A_1,\ldots,A_n$ (each of size $k\times m$) is {\bf proper} if
\be
    \sum_{i=1}^n A_i w_i = r
      \label{properA1}
\ee
for any vectors $w_i\in\IR^m$ and $r\in \IR^k$ such that
\be
   u_i^Tw_i + v_i^Tr = 0
      \label{properA2}
\ee
for all $1\leq i\leq n$. Then for any proper set of matrices
$A_1,\ldots,A_n$ the $k\times k$ matrix $D = \sum_{i=1}^n A_iA_i^T$ is
bounded from below by $B^{-1}$ in the sense that $D - B^{-1}$ is a
positive semidefinite matrix. The equality $D=B^{-1}$ holds if and only
if $A_i = - B^{-1} X_i$ for all $i=1,\ldots,n$.}
\\

This theorem is, probably, known, but we provide a full proof in
Appendix, for the sake of completeness.

As a direct consequence of the above theorem we obtain the lower
bound for our matrix ${\cal D}_1$:
\\

\noindent{\bf Theorem (Kanatani-Cramer-Rao lower bound)}. {\em We
have ${\cal D}_1\geq{\cal D}_{\min}$, in the sense that ${\cal
D}_1 - {\cal D}_{\min}$ is a positive semidefinite matrix, where}
\be
   {\cal D}_{\min}^{-1} = \sum_{i=1}^n
   \frac{(\nabla_{\Theta} P(\bar{\bf x}_i;\Theta))
   (\nabla_{\Theta} P(\bar{\bf x}_i;\Theta))^T}
   {\|\nabla_{{\bf x}}\, P(\bar{\bf x}_i;\Theta)\|^2}
     \label{Dmin}
\ee

In view of (\ref{Csig}) and (\ref{calC1}), the above theorem says that
the lower bound for the covariance matrix ${\cal C}_{\hat{\Theta}}$ is,
to the leading order,
\be
  {\cal C}_{\hat{\Theta}} \geq {\cal C}_{\min}
  = \sigma^2 {\cal D}_{\min}
    \label{RC}
\ee
The standard deviations of the components of the estimate
$\hat{\Theta}$ are of order $\sigma_{\hat{\Theta}} = {\cal
O}(\sigma)$. Therefore, the bias of $\hat{\Theta}$, which is at
most of order $\sigma^2$ by (\ref{Tbias}), is infinitesimally
small, as $\sigma \to 0$, compared to the standard deviations.
This means that the estimates satisfying (\ref{Tass}) are
practically unbiased.

The bound (\ref{RC}) was first derived by Kanatani
\cite{Ka96,Ka98} for the Cartesian functional model and strictly
unbiased estimates of $\Theta$, i.e.\ satisfying $E(\hat{\Theta})
=\bar{\Theta}$. One can easily derive (\ref{Tass}) from
$E(\hat{\Theta}) =\bar{\Theta}$ by taking the limit $\sigma \to
0$, hence our results generalize those of Kanatani.

\section{Statistical efficiency of algebraic fits}
\label{secSE} \setcounter{equation}{0}

Here we derive an explicit formula for the covariance matrix of the
weighted algebraic fit (\ref{Fmain3}) and describe the weights $w_i$
for which the fit is statistically efficient. For brevity, we write
$P_i = P(x_i,y_i;\Theta)$. We assume that the weight function
$w(x,y,;\Theta)$ is regular, in particular has bounded derivatives with
respect to $\Theta$, the next section will demonstrate the importance
of this condition. The solution of the minimization problem
(\ref{Fmain3}) satisfies
\be
   \sum P_i^2 \, \nabla_{\Theta} w_i +
   2 \sum w_i \, P_i \, \nabla_{\Theta} P_i = 0
      \label{weq}
\ee
Observe that $P_i = {\cal O} (\sigma)$, so that the first sum in
(\ref{weq}) is ${\cal O}(\sigma^2)$ and the second sum is ${\cal
O} (\sigma)$. Hence, to the leading order, the solution of
(\ref{weq}) can be found by discarding the first sum and solving
the reduced equation
\be
   \sum w_i\, P_i\, \nabla_{\Theta} P_i = 0
      \label{weq1}
\ee
More precisely, if $\hat{\Theta}_1$ and $\hat{\Theta}_2$ are
solutions of (\ref{weq}) and (\ref{weq1}), respectively, then
$\hat{\Theta}_1 -\bar{\Theta} = {\cal O} (\sigma)$,
$\hat{\Theta}_2 -\bar{\Theta} = {\cal O} (\sigma)$, and
$\|\hat{\Theta}_1 -\hat{\Theta}_2 \|= {\cal O} (\sigma^2)$.
Furthermore, the covariance matrices of $\hat{\Theta}_1$ and
$\hat{\Theta}_2$ coincide, to the leading order, i.e.\ ${\cal
C}_{\hat{\Theta}_1} {\cal C}_{\hat{\Theta}_2}^{-1} \to I$ as
$\sigma \to 0$. Therefore, in what follows, we only deal with the
solution of equation (\ref{weq1}).

To find the covariance matrix of $\hat{\Theta}$ satisfying
(\ref{weq1}) we put $\hat{\Theta} =\bar{\Theta} +\delta \Theta$
and ${\bf x}_i = \bar{\bf x}_i + \delta {\bf x}_i$ and obtain,
working to the leading order,
$$
  \sum w_i (\nabla_{\Theta} P_i)
  (\nabla_{\Theta} P_i)^T\, (\delta \Theta)
    = - \sum w_i (\nabla_{\bf x} P_i)^T \, (\delta {\bf x}_i) \,
    (\nabla_{\Theta} P_i) + {\cal O}(\sigma^2)
$$
hence
$$
   \delta \Theta  = -
   \left [ \sum w_i (\nabla_{\Theta} P_i)
   (\nabla_{\Theta} P_i)^T \right ]^{-1}
   \left [ \sum w_i (\nabla_{\bf x} P_i)^T \,
   (\delta {\bf x}_i)\, (\nabla_{\Theta} P_i)\right ]
    + {\cal O}(\sigma^2)
$$
The covariance matrix is then
\begin{eqnarray*}
   {\cal C}_{\hat{\Theta}} & = &
   E \left [ (\delta \Theta)\, (\delta \Theta)^T \right ]\\
   & = & \sigma^2
   \left [ \sum w_i (\nabla_{\Theta} P_i)
   (\nabla_{\Theta} P_i)^T \right ]^{-1}
   \left [ \sum w_i^2 \|\nabla_{\bf x} P_i\|^2
   (\nabla_{\Theta} P_i)
   (\nabla_{\Theta} P_i)^T \right ]\\
   & & \times \left [ \sum w_i (\nabla_{\Theta} P_i)
   (\nabla_{\Theta} P_i)^T \right ]^{-1}
   + {\cal O}(\sigma^3)
\end{eqnarray*}
Denote by ${\cal D}_2$ the principal factor here, i.e.\
$$
   {\cal D}_2 =
   \left [ \sum w_i (\nabla_{\Theta} P_i)
   (\nabla_{\Theta} P_i)^T \right ]^{-1}
   \left [ \sum w_i^2 \|\nabla_{\bf x} P_i\|^2
   (\nabla_{\Theta} P_i)
   (\nabla_{\Theta} P_i)^T \right ]\,
   \left [ \sum w_i (\nabla_{\Theta} P_i)
   (\nabla_{\Theta} P_i)^T \right ]^{-1}
$$
The following theorem establishes a lower bound for ${\cal D}_2$:
\\

\noindent{\bf Theorem}. {\em We have ${\cal D}_2\geq{\cal
D}_{\min}$, in the sense that ${\cal D}_2 - {\cal D}_{\min}$ is a
positive semidefinite matrix, where ${\cal D}_{\min}$ is given by
(\ref{Dmin}). The equality ${\cal D}_2 ={\cal D}_{\min}$ holds if
and only if $w_i = {\rm const}/\|\nabla_{{\bf x}}\, P_i\|^2$ for
all $i=1,\ldots,n$. In other words, an algebraic fit
(\ref{Fmain3}) is {\bf statistically efficient} if and only if the
weight function $w(x,y;\Theta)$ satisfies
\be
   w(x,y;\Theta) = \frac{c(\Theta)}{\|\nabla_{{\bf x}}\, P(x,y;\Theta)\|^2}
      \label{wopt}
\ee
for all triples $x,y,\Theta$ such that $P(x,y;\Theta)=0$. Here
$c(\Theta)$ may be an arbitrary function of $\Theta$.}
\\

The bound ${\cal D}_2\geq{\cal D}_{\min}$ here is a particular case of
the previous theorem. It also can be obtained directly from the linear
algebra theorem if one sets $u_i= \nabla_{\bf x} P_i$, $v_i=
\nabla_{\Theta} P_i$, and
$$
   A_i = - w_i\, \left [ \sum_{j=1}^n w_j (\nabla_{\Theta} P_j)
   (\nabla_{\Theta} P_j)^T \right ]^{-1}
   (\nabla_{\Theta} P_i) \,
   (\nabla_{\bf x} P_i)^T
$$
for $1\leq i\leq n$.

The expression (\ref{wopt}) characterizing the efficiency, follows from
the last claim in the linear algebra theorem.

\section{Circle fit}
\label{secCF} \setcounter{equation}{0}

Here we illustrate our conclusions by the relatively simple
problem of fitting circles. The canonical equation of a circle is
\be
         (x-a)^2+ (y-b)^2 -R^2=0
           \label{circ0}
\ee
and we need to estimate three parameters $a,b,R$. The simple
algebraic fit (\ref{Fmain2}) takes form
\be
       {\cal F}_2(a,b,R) =
       \sum_{i=1}^n [(x_i-a)^2+ (y_i-b)^2 -R^2]^2
       \ \  \to\ \  \min
         \label{F2}
\ee
and the weighted algebraic fit (\ref{Fmain3}) takes form
\be
       {\cal F}_3(a,b,R) =
       \sum_{i=1}^n w_i [(x_i-a)^2+ (y_i-b)^2 -R^2]^2
       \ \  \to\ \  \min
         \label{F3}
\ee
In particular, the GRAF becomes
\be
       {\cal F}_4(a,b,R) =
       \sum_{i=1}^n \frac{[(x_i-a)^2+ (y_i-b)^2 -R^2]^2}
       {(x_i-a)^2+ (y_i-b)^2}
       \ \  \to\ \  \min
         \label{F4}
\ee
(where the irrelevant constant factor of 4 in the denominator is
dropped).

In terms of (\ref{Dmin}), we have
$$
  \nabla_{\Theta} P(\bar{\bf x}_i;\Theta)
  = -2(\bar{x}_i-a,\bar{y}_i-b,R)^T
$$
and $\nabla_{{\bf x}}\, P(\bar{\bf x}_i;\Theta) =
2(\bar{x}_i-a,\bar{y}_i-b)^T$, hence
$$
  \|\nabla_{{\bf x}}\, P(\bar{\bf x}_i;\Theta)\|^2 =
    4[(\bar{x}_i-a)^2+(\bar{y}_i-b)^2]=4R^2
$$
Therefore,
\be
    {\cal D}_{\min} = \left (\begin{array}{ccc}
    \sum u_i^2 & \sum u_iv_i & \sum u_i \\
    \sum u_iv_i & \sum v_i^2 & \sum v_i \\
    \sum u_i & \sum v_i & n \\
    \end{array} \right )^{-1}
      \label{Dmincir}
\ee
where we denote, for brevity,
$$
   u_i=\frac{\bar{x}_i-a}{R},\ \ \ \
   v_i=\frac{\bar{y}_i-b}{R}
$$
The above expression for ${\cal D}_{\min}$ was derived earlier in
\cite{CT95,Ka98}.

Now, our Theorem in Section~\ref{secSE} shows that the weighted
algebraic fit (\ref{F3}) is statistically efficient if and only if
the weight function satisfies $w(x,y;a,b,R)=c(a,b,R)/(4R^2)$.
Since $c(a,b,R)$ may be an arbitrary function, then the
denominator $4R^2$ here is irrelevant. Hence, statistically
efficiency is achieved whenever $w(x,y;a,b,R)$ is simply
independent of $x$ and $y$ for all $(x,y)$ lying on the circle. In
particular, the GRAF (\ref{F4}) is statistically efficient because
$w(x,y;a,b,R)=[(x-a)^2+(y-b)^2]^{-1}=R^{-2}$. The simple AF
(\ref{F2}) is also statistically efficient since $w(x,y;a,b,R)=1$.

We note that the GRAF (\ref{F4}) is a highly nonlinear problem, and in
its exact form (\ref{F4}) is not used in practice. Instead, there are
two modifications of GRAF popular among experimenters. One is due to
Chernov and Ososkov \cite{CO84} and Pratt \cite{Pr87}:
\be
       {\cal F}_4'(a,b,R) =
       R^{-2}\sum_{i=1}^n [(x_i-a)^2+ (y_i-b)^2 -R^2]^2
       \ \  \to\ \  \min
         \label{F4a}
\ee
(it is based on the approximation $(x_i-a)^2+ (y_i-b)^2 \approx R^2$),
and the other due to Agin \cite{Ag81} and Taubin \cite{Ta91}:
\be
       {\cal F}_4''(a,b,R) =
       \frac{1}{\sum (x_i-a)^2+ (y_i-b)^2}
       \sum_{i=1}^n [(x_i-a)^2+ (y_i-b)^2 -R^2]^2
       \ \  \to\ \  \min
         \label{F4b}
\ee
(here one simply averages the denominator of (\ref{F4}) over $1\leq
i\leq n$). We refer the reader to \cite{CL02} for a detailed analysis
of these and other circle fitting algorithms, including their numerical
implementations.

We have tested experimentally the efficiency of four circle fitting
algorithms: the OLSF (\ref{Fmain1}), the simple AF (\ref{F2}), the
Pratt method (\ref{F4a}), and the Taubin method (\ref{F4b}). We have
generated $n=20$ points equally spaced on a circle, added an isotropic
Gaussian noise with variance $\sigma^2$ (according to the Cartesian
model), and estimated the efficiency of the estimate of the center by
\be
   E = \frac{\sigma^2 ({\cal D}_{11}+{\cal D}_{22})}
   {\la (\hat{a}-a)^2 + (\hat{b}-b)^2 \ra}
      \label{E}
\ee
Here $(a,b)$ is the true center, $(\hat{a},\hat{b})$ is its estimate,
$\la \cdots \ra$ denotes averaging over many random samples, and ${\cal
D}_{11}$, ${\cal D}_{22}$ are the first two diagonal entries of the
matrix (\ref{Dmincir}). Table~1 shows the efficiency of the above
mentioned four algorithms for various values of $\sigma/R$. We see that
they all perform very well, and indeed are efficient as $\sigma\to 0$.
One might notice that the OLSF slightly outperforms the other methods,
and the AF is the second best.

\begin{center}
\begin{tabular}{||r||c|c|c|c||}
\hline\hline $\sigma/R$ & OLSF & AF & Pratt & Taubin \\
\hline \hline $<0.01$ & $\sim 1$ & $\sim 1$ & $\sim 1$ & $\sim 1$ \\
        \hline 0.01 & 0.999 & 0.999 & 0.999 & 0.999 \\ \hline
       0.02 & 0.999 & 0.998 & 0.997 & 0.997 \\ \hline
       0.03 & 0.998 & 0.996 & 0.995 & 0.995 \\ \hline
       0.05 & 0.996 & 0.992 & 0.987 & 0.987 \\ \hline
       0.10 & 0.985 & 0.970 & 0.953 & 0.953 \\ \hline
       0.20 & 0.935 & 0.900 & 0.837 & 0.835 \\ \hline
       0.30 & 0.825 & 0.824 & 0.701 & 0.692 \\ \hline
\hline
\end{tabular}\vspace*{0.2cm}
\end{center}

\begin{center}
Table 1. Efficiency of circle fitting algorithms. Data are sampled
along a full circle.
\end{center}

Table~2 shows the efficiency of the same algorithms as the data points
are sampled along half a circle, rather than a full circle. Again, the
efficiency as $\sigma\to 0$ is clear, but we also make another
observation. The AF now consistently falls behind the other methods for
all $\sigma/R\leq 0.2$, but for $\sigma/R=0.3$ the others suddenly
break down, while the AF keeps afloat.

\begin{center}
\begin{tabular}{||r||c|c|c|c||}
\hline\hline $\sigma/R$ & OLSF & AF & Pratt & Taubin \\
\hline \hline $<0.01$ & $\sim 1$ & $\sim 1$ & $\sim 1$ & $\sim 1$ \\
        \hline 0.01 & 0.999 & 0.996 & 0.999 & 0.999 \\ \hline
       0.02 & 0.997 & 0.983 & 0.997 & 0.997 \\ \hline
       0.03 & 0.994 & 0.961 & 0.992 & 0.992 \\ \hline
       0.05 & 0.984 & 0.902 & 0.978 & 0.978 \\ \hline
       0.10 & 0.935 & 0.720 & 0.916 & 0.916 \\ \hline
       0.20 & 0.720 & 0.493 & 0.703 & 0.691 \\ \hline
       0.30 & 0.122 & 0.437 & 0.186 & 0.141 \\ \hline
\hline
\end{tabular}\vspace*{0.2cm}
\end{center}

\begin{center}
Table 2. Efficiency of circle fitting algorithms with data sampled
along half a circle.
\end{center}

The reason of the above turnaround is that at large noise the data
points may occasionally line up along a circular arc of a very large
radius. Then the OLSF, Pratt and Taubin dutifully return a large circle
whose center lies far away, and such fits blow up the denominator of
(\ref{E}), a typical effect of large outliers. On the contrary, the AF
is notoriously known for its systematic bias toward smaller circles
\cite{CO84,GGS94,Pr87}, hence while it is less accurate than other fits
for typical random samples, its bias safeguards it from large outliers.

This behavior is even more pronounced when the data are sampled along
quarter\footnote{All our algorithms are invariant under simple
geometric transformations such as translations, rotations and
similarities, hence our experimental results do not depend on the
choice of the circle, its size, and the part of the circle the data are
sampled from.} of a circle (Table~3). We see that the AF is now far
worse than the other fits for $\sigma/R<0.1$ but the others
characteristically break down at some point ($\sigma/R=0.1$).

\begin{center}
\begin{tabular}{||r||c|c|c|c||}
\hline\hline $\sigma/R$ & OLSF & AF & Pratt & Taubin \\
\hline \hline  0.01 & 0.997 & 0.911 & 0.997 & 0.997 \\ \hline
       0.02 & 0.977 & 0.722 & 0.978 & 0.978 \\ \hline
       0.03 & 0.944 & 0.555 & 0.946 & 0.946 \\ \hline
       0.05 & 0.837 & 0.365 & 0.843 & 0.842 \\ \hline
       0.10 & 0.155 & 0.275 & 0.163 & 0.158 \\ \hline
\hline
\end{tabular}\vspace*{0.2cm}
\end{center}

\begin{center}
Table 3. Data are sampled along a quarter of a circle.
\end{center}

It is interesting to test smaller circular arcs, too. Figure 1
shows a color-coded diagram of the efficiency of the OLSF and the
AF for arcs from $0^{\rm o}$ to $50^{\rm o}$ and variable $\sigma$
(we set $\sigma=ch$, where $h$ is the height of the circular arc,
see Fig.~2, and $c$ varies from 0 to 0.5). The efficiency of the
Pratt and Taubin is virtually identical to that of the OLSF, so it
is not shown here. We see that the OLSF and AF are efficient as
$\sigma\to 0$ (both squares in the diagram get white at the
bottom), but the AF loses its efficiency at moderate levels of
noise ($c>0.1$), while the OLSF remains accurate up to $c=0.3$
after which it rather sharply breaks down.

\vspace*{10mm} \centerline{\epsffile{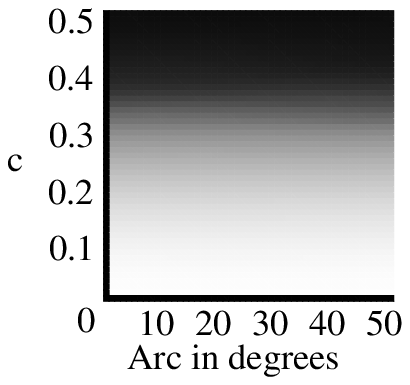}$\ \ \ \ $
\epsffile{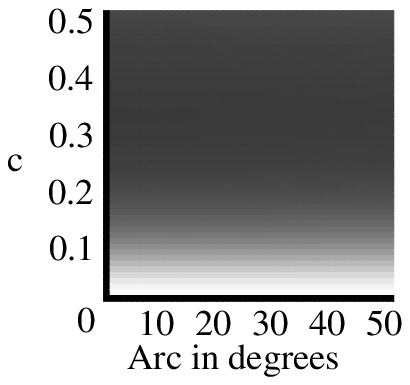} $\ \ \ \ $ \epsffile{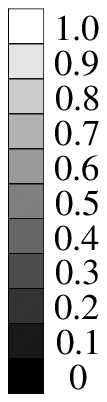}}

\begin{center}
Figure 1: The efficiency of the simple OLSF (left) and the AF (center).
The bar on the right explains color codes.
\end{center} \vspace*{5mm}

The following analysis sheds more light on the behavior of the
circle fitting algorithms. When the curvature of the arc
decreases, the center coordinates $a,b$ and the radius $R$ grow to
infinity and their estimates become highly unreliable. In that
case the circle equation (\ref{circ0}) can be converted to a more
convenient algebraic form
\be
     A(x^2+y^2) + Bx + Cy + D = 0
       \label{ABCD}
\ee
with an additional constrain on the parameters: $B^2+C^2-4AD = 1$. This
parametrization was used in \cite{Pr87,GGS94}, and analyzed in detail
in \cite{CL02}. We note that the original parameters can be recovered
via $a=-B/2A$, $b=-C/2A$, and $R=(2\,|A|)^{-1}$. The new
parametrization (\ref{ABCD}) is safe to use for arcs with arbitrary
small curvature: the parameters $A,B,C,D$ remain bounded and never
develop singularities, see \cite{CL02}. Even as the curvature vanishes,
we simply get $A=0$, and the equation (\ref{ABCD}) represents a line
$Bx+Cy+D=0$.

\vspace*{5mm} \centerline{\epsffile{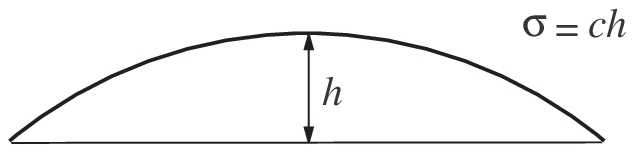}}

\begin{center}
Figure 2: The height of an arc, $h$, and our formula for $\sigma$.
\end{center} \vspace*{5mm}

In terms of the new parameters $A,B,C,D$, the weighted algebraic fit
(\ref{Fmain3}) takes form
\be
       {\cal F}_3(A,B,C,D) =
       \sum_{i=1}^n w_i [A(x^2+y^2) + Bx + Cy + D]^2
       \ \  \to\ \  \min
         \label{FF3}
\ee
(under the constraint $B^2+C^2-4AD = 1$). Converting the AF (\ref{F2})
to the new parameters gives
\be
       {\cal F}_2(A,B,C,D) =
       \sum_{i=1}^n A^{-2} [A(x^2+y^2) + Bx + Cy + D]^2
       \ \  \to\ \  \min
         \label{FF2}
\ee
which corresponds to the weight function $w=1/A^2$. The Pratt method
(\ref{F4a}) turns to
\be
       {\cal F}_4(A,B,C,D) =
       \sum_{i=1}^n [A(x^2+y^2) + Bx + Cy + D]^2
       \ \  \to\ \  \min
         \label{FF4}
\ee
We now see why the AF is unstable and inaccurate for arcs with
small curvature: its weight function $w=1/A^2$ develops a
singularity (it explodes) in the limit $A\to 0$. Recall that, in
our derivation of the statistical efficiency theorem (Section~3),
we assumed that the weight function was regular (had bounded
derivatives). This assumption is clearly violated by the AF
(\ref{FF2}). On the contrary, the Pratt fit (\ref{FF4}) uses a
safe choice $w=1$ and thus behaves decently on arcs with small
curvature, see next.

\vspace*{10mm} \centerline{\epsffile{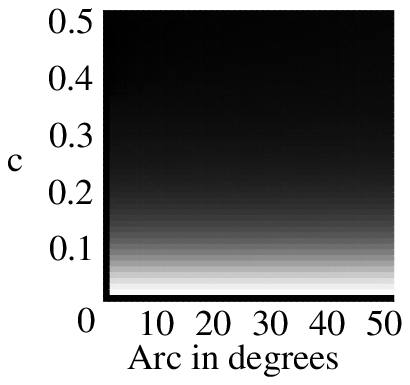}$\ \ \ \ $
\epsffile{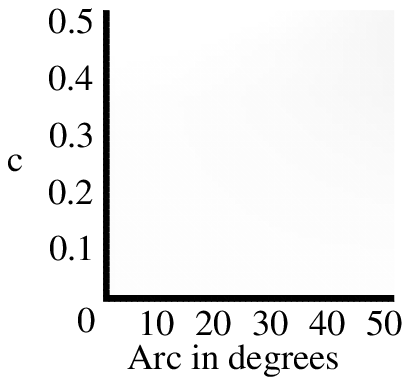} $\ \ \ \ $ \epsffile{Bar.eps}}

\begin{center}
Figure 3: The efficiency of the simple AF (left) and the Pratt
method (center). The bar on the right explains color codes.
\end{center} \vspace*{5mm}

Figure 3 shows a color-coded diagram of the efficiency of the estimate
of the parameter\footnote{Note that $|A|=1/2R$, hence the estimation of
$A$ is equivalent to that of the curvature, an important geometric
parameter of the arc.} $A$ by the AF (\ref{FF2}) versus Pratt
(\ref{FF4}) for arcs from $0^{\rm o}$ to $50^{\rm o}$ and the noise
level $\sigma=ch$, where $h$ is the height of the circular arc and $c$
varies from 0 to 0.5. The efficiency of the OLSF and the Taubin method
is visually indistinguishable from that of Pratt (the central square in
Fig.~3), so we did not include it here.

We see that the AF performs significantly worse than the Pratt
method for all arcs and most of the values of $c$ (i.e.,
$\sigma$). The Pratt's efficiency is close 100\%, its lowest point
is 89\% for $50^{\rm o}$ arcs and $c=0.5$ (the top right corner of
the central square barely gets grey). The AF's efficiency is below
10\% for all $c>0.2$ and almost zero for $c>0.4$. Still, the AF
remains efficient as $\sigma\to 0$ (as the tiny white strip at the
bottom of the left square proves), but its efficiency can be only
counted on when $\sigma$ is extremely small.

Our analysis demonstrates that the choice of the weights $w_i$ in the
weighted algebraic fit (\ref{Fmain3}) should be made according to our
theorem in Section~3, and, in addition, one should avoid singularities
in the domain of parameters.

\renewcommand{\theequation}{A.\arabic{equation}}

\section*{Appendix}
\label{secA} \setcounter{equation}{0}

Here we prove the theorem of linear algebra stated in
Section~\ref{secKCR}. For the sake of clarity, we divide our proof into
small lemmas:
\medskip

\noindent{\bf Lemma 1}. {\em The matrix $B$ is indeed nonsingular}.

{\em Proof}. If $Bz=0$ for some nonzero vector $z\in \IR^k$, then
$0 = z^TBz = \sum_{i=1}^n (v_i^Tz)^2/\|u_i\|^2$, hence $v_i^Tz=0$
for all $1\leq i\leq k$, a contradiction.
\medskip

\noindent{\bf Lemma 2}. {\em If a set of $n$ matrices $A_1,\ldots,A_n$
is proper, then rank$(A_i)\leq 1$. Furthermore, each $A_i$ is given by
$A_i = z_iu_i^T$ for some vector $z_i\in \IR^k$, and the vectors
$z_1,\ldots,z_n$ satisfy $\sum_{i=1}^n z_iv_i^T = -I$ where $I$ is the
$k\times k$ identity matrix. The converse is also true.}

{\em Proof}. Let vectors $w_1,\ldots,w_n$ and $r$ satisfy the
requirements (\ref{properA1}) and (\ref{properA2}) of the theorem.
Consider the orthogonal decomposition $w_i = c_iu_i + w_i^\perp$ where
$w_i^\perp$ is perpendicular to $u_i$, i.e.\ $u_i^Tw_i^\perp = 0$. Then
the constraint (\ref{properA2}) can be rewritten as
\be
    c_i = -\frac{v_i^Tr}{u_i^Tu_i}
      \label{properA3}
\ee
for all $i=1,\ldots,n$ and (\ref{properA1}) takes form
\be
    \sum_{i=1}^n c_iA_iu_i + \sum_{i=1}^n A_iw_i^\perp = r
      \label{properA4}
\ee
We conclude that $A_iw_i^\perp = 0$ for every vector $w_i^\perp$
orthogonal to $u_i$, hence $A_i$ has a $(k-1)$-dimensional kernel, so
indeed its rank is zero or one. If we denote $z_i = A_iu_i/ \|u_i\|^2$,
we obtain $A_i=z_iu_i^T$. Combining this with
(\ref{properA3})-(\ref{properA4}) gives
$$
   r = - \sum_{i=1}^n (v_i^Tr)z_i =
   - \left (\sum_{i=1}^n z_iv_i^T\right )\, r
$$
Since this identity holds for any vector $r\in \IR^k$, the expression
within parentheses is $-I$. The converse is obtained by straightforward
calculations. Lemma is proved. \medskip

\noindent{\bf Corollary}. {\em Let ${\bf n}_i = u_i/\|u_i\|$. Then
$A_i{\bf n}_i{\bf n}_i^TA_i = A_iA_i^T$ for each $i$}.
\medskip

This corollary implies our lemma stated in Section~\ref{secKCR}. We now
continue the proof of the theorem.\medskip

\noindent{\bf Lemma 3}. {\em The sets of proper matrices make a linear
variety, in the following sense. Let $A_1',\ldots,A_n'$ and
$A_1'',\ldots,A_n''$ be two proper sets of matrices, then the set
$A_1,\ldots,A_n$ defined by $A_i = A_i' + c(A_i''- A_i')$ is proper for
every $c\in\IR$.}

{\em Proof}. According to the previous lemma, $A_i'=z_i'u_i^T$ and
$A_i''=z_i''u_i^T$ for some vectors $z_i',z_i''$, $1\leq i\leq n$.
Therefore, $A_i=z_iu_i^T$ for $z_i= z_i' + c(z_i''- z_i')$. Lastly,
$$
   \sum_{i=1}^n z_iv_i^T = \sum_{i=1}^n z_i'v_i^T
   +c\sum_{i=1}^n z_i''v_i^T - c\sum_{i=1}^n z_i'v_i^T = - I
$$
Lemma is proved.
\medskip

\noindent{\bf Lemma 4}. {\em If a set of $n$ matrices $A_1,\ldots,A_n$
is proper, then $\sum_{i=1}^n A_iX_i^T = -I$, where $I$ is the $k\times
k$ identity matrix.}

{\em Proof}. By using Lemma~2 $\sum_{i=1}^n A_iX_i^T =
\sum_{i=1}^n z_iv_i^T = -I$. Lemma is proved.
\medskip

\noindent{\bf Lemma 5}. {\em We have indeed $D \geq B^{-1}$.}

{\em Proof}. For each $i=1,\ldots,n$ consider the $2k\times m$
matrix $Y_i = \left (\begin{array}{c} A_i\\X_i\end{array} \right
)$. Using the previous lemma gives
$$
   \sum_{i=1}^n Y_i\,Y_i^T =
   \left (\begin{array}{rr} D & -I\\ -I & B
   \end{array} \right )
$$
By construction, this matrix is positive semidefinite. Hence, the
following matrix is also positive semidefinite:
$$
   \left (\begin{array}{rr} I & B^{-1} \\ 0 & B^{-1}
   \end{array} \right )
   \left (\begin{array}{rr} D & -I\\ -I & B
   \end{array} \right )
   \left (\begin{array}{cc} I & 0 \\ B^{-1} & B^{-1}
   \end{array} \right ) =
   \left (\begin{array}{cc} D-B^{-1} & 0\\ 0 & B^{-1}
   \end{array} \right )
$$
By Sylvester's theorem, the matrix $D-B^{-1}$ is positive semidefinite.
\medskip

\noindent{\bf Lemma 6}. {\em The set of matrices $A_i^{\rm o} = -
B^{-1} X_i$ is proper, and for this set we have $D=B^{-1}$.}

{\em Proof}. Straightforward calculation.
\medskip

\noindent{\bf Lemma 7}. {\em If $D=B^{-1}$ for some proper set of
matrices $A_1,\ldots,A_n$, then $A_i=A_i^{\rm o}$ for all $1\leq i\leq
n$}.

{\em Proof}. Assume that there is a proper set of matrices
$A_1',\ldots,A_n'$, different from $A_1^{\rm o},\ldots,A_n^{\rm
o}$, for which $D=B^{-1}$. Denote $\delta A_i = A_i'-A_i^{\rm o}$.
By Lemma 3, the set of matrices $A_i(\gamma) = A_i^{\rm o} +
\gamma (\delta A_i)$ is proper for every real $\gamma$. Consider
the variable matrix
\begin{eqnarray*}
   D(\gamma) & = & \sum_{i=1}^n [A_i(\gamma)] [A_i(\gamma)]^T\\
   & = & \sum_{i=1}^n A_i^{\rm o}(A_i^{\rm o})^T
   + \gamma\left (\sum_{i=1}^n A_i^{\rm o}(\delta A_i)^T
   +\sum_{i=1}^n (\delta A_i)(A_i^{\rm o})^T\right )
   +\gamma^2\sum_{i=1}^n (\delta A_i)(\delta A_i)^T
\end{eqnarray*}
Note that the matrix $R = \sum_{i=1}^n A_i^{\rm o}(\delta A_i)^T
+\sum_{i=1}^n (\delta A_i)(A_i^{\rm o})^T$ is symmetric. By
Lemma~5 we have $D(\gamma)\geq B^{-1}$ for all $\gamma$, and by
Lemma~6 we have $D(0)=B^{-1}$. It is then easy to derive that
$R=0$. Next, the matrix $S =\sum_{i=1}^n (\delta A_i)(\delta
A_i)^T$ is symmetric positive semidefinite. Since we assumed that
$D(1)=D(0)=B^{-1}$, it is easy to derive that $S=0$ as well.
Therefore, $\delta A_i = 0$ for every $i=1,\ldots,n$. The theorem
is proved.

\end{document}